\documentclass{article}
\usepackage{ijcai24}

\usepackage{times}
\usepackage{soul}
\usepackage{url}
\usepackage[hidelinks]{hyperref}
\usepackage[utf8]{inputenc}
\usepackage[small]{caption}
\usepackage{graphicx}
\usepackage{amsmath}
\usepackage{amsthm}
\usepackage{booktabs}
\usepackage{algorithm}
\usepackage{algorithmic}
\usepackage[switch]{lineno}
\usepackage{color}
\usepackage{amsfonts}
\usepackage{subcaption}
\usepackage[sectionbib,square]{natbib}

\usepackage{multirow}
\usepackage{multicol}

\newcommand{\cut}[1]{}

\title{Large Language Model as a Policy Teacher for Training \\ Reinforcement Learning Agents}

\author{
Zihao Zhou\and
Bin Hu\and
Chenyang Zhao\and
Pu Zhang\And
Bin Liu\footnote{Corresponding author (Email:bins@ieee.org)}
\affiliations
Zhejiang Lab\\
\emails
\{zhouzihao,hubin,c.zhao,puz,liubin\}@zhejianglab.com
}
\begin{document}

\maketitle

\begin{abstract}
    Recent studies have uncovered the potential of Large Language Models (LLMs) in addressing complex sequential decision-making tasks through the provision of high-level instructions. However, LLM-based agents lack specialization in tackling specific target problems, particularly in real-time dynamic environments. Additionally, deploying an LLM-based agent in practical scenarios can be both costly and time-consuming. On the other hand, reinforcement learning (RL) approaches train agents that specialize in the target task but often suffer from low sampling efficiency and high exploration costs. In this paper, we introduce a novel framework that addresses these challenges by training a smaller, specialized student RL agent using instructions from an LLM-based teacher agent. By incorporating the guidance from the teacher agent, the student agent can distill the prior knowledge of the LLM into its own model. Consequently, the student agent can be trained with significantly less data. Moreover, through further training with environment feedback, the student agent surpasses the capabilities of its teacher for completing the target task. We conducted experiments on challenging MiniGrid and Habitat environments, specifically designed for embodied AI research, to evaluate the effectiveness of our framework. The results clearly demonstrate that our approach achieves superior performance compared to strong baseline methods. Our code is available at https://github.com/ZJLAB-AMMI/LLM4Teach.
\end{abstract}

\section{Introduction}

Large Language Models (LLMs) have revolutionized the field of artificial intelligence. These models are trained with an internet-scale text corpus, enabling them to exhibit remarkable capabilities such as natural language generation, question answering, and translation \citep{brown2020language,du2022glm,vicuna2023}.
Previous work suggests that these models contain vast general knowledge about the world and are capable of solving complex reasoning problems \citep{radford2019language,brown2020language,wei2022chain}. Recently, several works have attempted to use LLMs to generate action plans in an embodied environment \citep{ahn2022can,wang2023voyager,driess2023palm,song2023llm,sha2023languagempc,mao2023gpt}.
However, LLMs face challenges in generating effective end-to-end instructions for specific embodied tasks, especially in real-world dynamic scenarios. This limitation arises from two key factors. Firstly, LLMs do not possess the appropriate task incentives during the training process. Secondly, these models lack the capability to actively interact with the environment and gather real-time data \citep{carta2023grounding}. Furthermore, the utilization of LLMs often requires substantial computational resources, e.g.,  memory and power. These requirements render their deployment impractical and expensive, especially when considering their use on lightweight edge devices. These challenges motivate us to address the following question:

\textit{How do we develop a lightweight, specialized agent that can quickly acquire the capabilities of LLMs for a specific sequential decision-making task? }

A commonly used solution is to train a specialized reinforcement learning (RL) based agent that starts learning from scratch. However, this approach often incurs a significant exploration cost, especially in high-dimensional and complex embodied environments with sparse reward signals, due to the low sampling efficiency of RL methods.

In this paper, we propose a novel approach called \textit{LLM for policy teaching} (LLM4Teach), which utilizes a pre-trained LLM to expedite the training process of a small-scale RL-based student agent specialized for a target task. Specifically, in the early stage of training, the student agent queries the LLM-based teacher agent for action instructions and learns to mimic the behavior of its teacher through minimizing a distillation loss. As the learning process proceeds, the student gradually shifts from learning from its teacher to learning from the environment by upweighting a conventional RL loss. In another word, the objective function used for policy training is defined as a weighted average of the distillation loss and the RL loss. Since it allows the student agent to not only incorporate guidance from its LLM teacher but also learn from online interactions with the environment, LLM4Teach enables the student agent to identify and correct any mistakes made by its teacher, leading to improved performance on the target task compared to its teacher. Note that only the student agent is deployed and it shall not interact with the LLM in the test phase. That means the model finally deployed is very lightweight compared to an LLM.

To summarize, our \textbf{\textit{main contributions}} are:
\begin{itemize}
    \item We propose LLM4Teach, a policy distillation approach to address the limitations of LLM and RL-based agents for embodied sequential decision making.
    \item We demonstrate the performance of our approach empirically by extensive experiments conducted on challenging embodied environments. In contrast to LLM-based agents, our approach shows improved accuracy and decreased computational workload. In comparison to RL-based agents, it has much greater sample efficiency.
    \item As a byproduct, we demonstrate that relying solely on LLM can result in various types of incorrect decisions in embodied settings, while LLM4Teach offers an effective approach to mitigate or avoid the influence caused by such incorrect decisions. We also verify that offering uncertainty-aware rather than deterministic guidance through LLM can improve the sample efficiency for the student agent.
\end{itemize}

\section{Related Work}

In this paper, we consider an algorithmic agent operating in an open dynamic environment. The agent is required to make a series of decisions and take actions based on the current state of the environment, employing a specific policy to successfully complete a designated task. Here we provide a brief overview of relevant research in the literature.

\subsection{LLM-based Agents}

LLMs have exhibited impressive reasoning abilities, motivating researchers to employ them as fundamental components for constructing LLM-based agents in diverse decision-making scenarios \citep{xi2023rise,yang2023foundation,wang2023survey,biggie2023tell,zhen2023robot}. Recent research has demonstrated that LLMs can generate high-level plans in response to natural language descriptions of a given situation \citep{huang2022language,shinn2023reflexion,yao2022react} . However, these plans may propose actions that are not compatible with the acting agent or the environment due to a lack of grounding in the specific problem domain. In addressing this issue, \citet{ahn2022can} proposed grounding LLMs through an affordance function of pre-trained skills, which assists LLMs in formulating feasible plans for execution by the agents. Additionally, \citet{carta2023grounding} proposed an approach in which the agent interacts with the environment and subsequently fine-tunes the LLMs using online collected data, thereby enhancing adaptation to the target task. However, frequent interaction with an LLM can be costly. Therefore, \citet{hu2024enable} suggested an intelligent interaction approach that employs RL to determine when it is necessary to query the LLM, thus avoiding unnecessary interactions. Furthermore, \citet{nottingham2023selective} optimized the selection of information presented to LLMs, thereby reducing the length of input contexts. While these methods reduce the cost of utilizing LLMs for decision-making tasks, they all necessitate online access to a pre-trained LLM when the agent is deployed online during the testing phase.

In contrast, our approach involves utilizing the LLM solely during the training phase to distill task-specific knowledge from the LLM into a RL-based agent. Subsequently, during the testing phase, only the lightweight student agent is deployed, which works independently without dependence on the LLM.

\begin{figure*}[t]
    \centering
        \includegraphics[width= 0.8\textwidth]{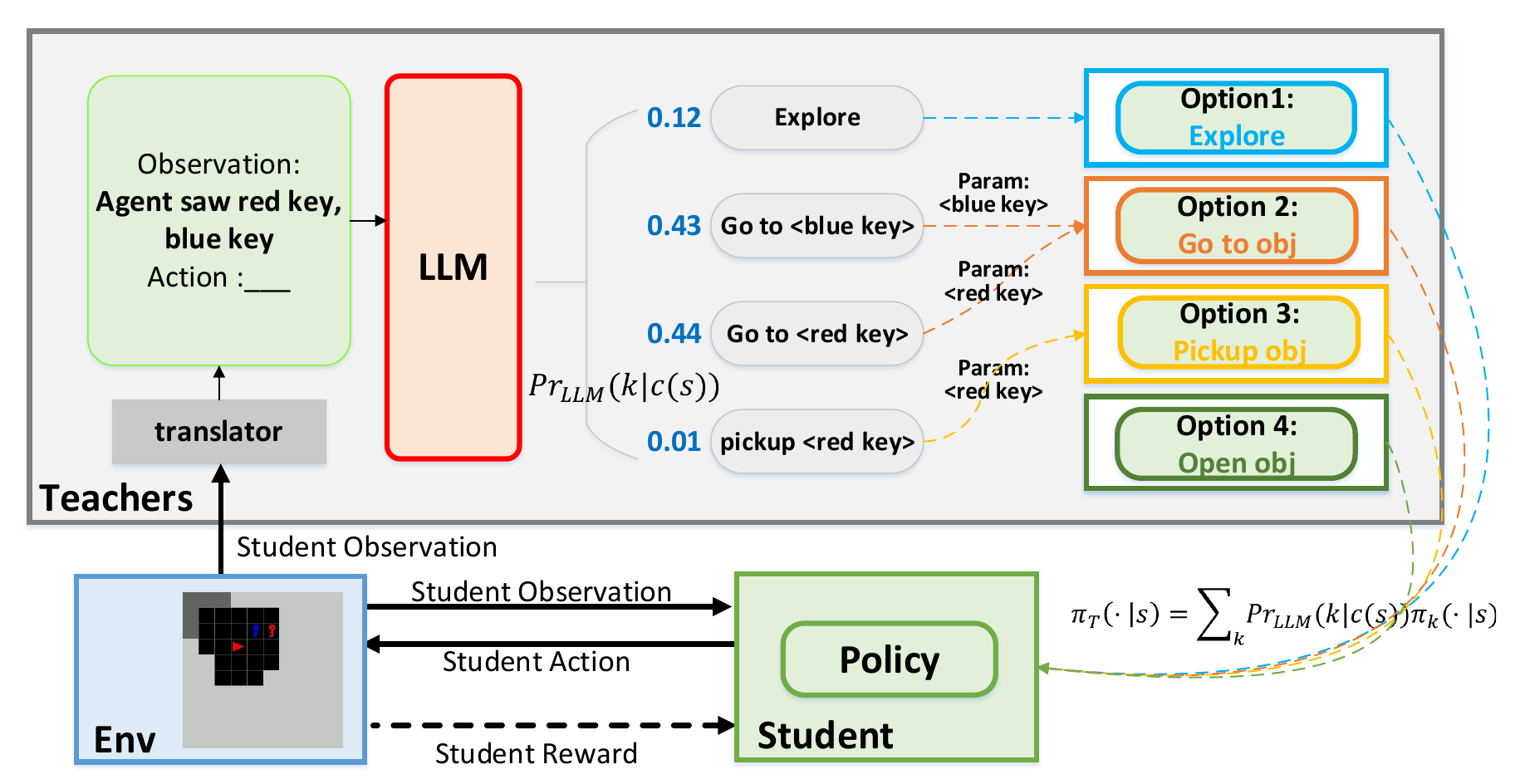}
        \caption{An illustration of our LLM4Teach framework using the MiniGrid environment as an exemplar. The LLM-based teacher agent responds to observations of the state provided by the environment by offering soft instructions. These instructions take the form of a distribution over a set of suggested actions. The student agent is trained to optimize two objectives simultaneously. The first one is to maximize the expected return, the same as in traditional RL algorithms. The other one is to encourage the student agent to follow the guidance provided by the teacher. As the student agent's expertise increases during the training process, the weight assigned to the second objective gradually decreases over time, reducing its reliance on the teacher.}
        \label{fig:framework}
\end{figure*}

\subsection{LLM Assisted RL}
Several studies have investigated the potential of utilizing LLMs to support the standard RL process by tapping into the general knowledge embedded in LLMs. For example,  \cite{kwon2023reward,yu2023language} and \cite{klissarov2023motif} employ LLMs to assist in assigning rewards. \cite{kwon2023reward} use LLMs as proxy reward functions to automatically label trajectory data with rewards, while \cite{yu2023language} utilize LLMs to flexibly define reward parameters for optimizing and completing various robot tasks. In a different approach, \cite{klissarov2023motif} leverage an offline dataset of behaviors and use LLMs' preferences over pairs of randomly sampled trajectories to construct a reward model. Furthermore, \cite{du2023guiding} and \cite{colas2023augmenting} focus on learning diverse behaviors without relying on reward supervision, employing LLMs to generate novel goals during exploration in the environment.
In contrast to these previous works, our approach focuses on leveraging prior knowledge about the target task to enhance the initial exploration stage of an RL agent. This allows us to train the policy model with significantly less data, thereby improving the efficiency of the learning process.

\subsection{Learning from Teacher Agents}
Prior research has sought to improve the inefficiencies of tabula rasa RL by utilizing existing teacher agents to guide the learning process of a specialized RL agent for a specific target problem \citep{da2020agents,agarwal2022reincarnating}. These instructions can manifest as demonstrations \citep{schaal1996learning}, scalar feedback \citep{knox2009interactively}, or action advice \citep{da2017simultaneously}. Jump-start RL involves the use of a teacher agent to assist in gathering high-quality data during the initial exploration phase of RL \citep{uchendu2023jump}. Kickstarting RL combines on-policy distillation with RL, prompting the student agent to emulate the teacher's behavior while optimizing for accumulated returns \citep{schmitt2018kickstarting}. \cite{matthews2022hierarchical} further extends this approach to hierarchical policies, transferring pre-trained low-level skill policies as teachers and training the student agent alongside a policy-over-teachers from scratch, which weighs the advice from each teacher agent at every time step.

In contrast, our approach does not rely on specialized teacher agents for the target problem. Instead, we harness the extensive general knowledge embedded in LLMs to expedite the learning process of the student agent through on-policy distillation for combining pre-trained fundamental skills to complete the target task.

\section{LLM4Teach}

In this section, we present our methodology LLM4Teach. To begin with, we fix the notations as follows.

We consider a sequential decision-making problem formalized as a Markov Decision Process (MDP), denoted by $\langle \mathcal{S}, \mathcal{A}, \mathcal{T}, \mathcal{R}, \gamma \rangle$, where $\mathcal{S}$ and $\mathcal{A}$ denote the state and action spaces, respectively. The transition probability function is denoted as $\mathcal{T}:\mathcal{S}\times \mathcal{A} \rightarrow \mathcal{P}(\mathcal{S})$, and the reward function is denoted as $\mathcal{R}: \mathcal{S} \times \mathcal{A} \times \mathcal{S} \rightarrow \mathbb{R}$. Additionally, $\gamma$ represents the discount factor. The primary objective is to learn an optimal policy $\pi:\mathcal{S} \rightarrow \mathcal{P}(\mathcal{A})$, which maximizes the expected cumulative return over time: $\max_{\pi} \mathbb{E} [\sum_t \gamma^t r_t]$. The parameter of the policy $\pi$ is denoted as $\theta$. A standard gradient-based RL algorithm minimizes a surrogate loss, $\mathcal{L}_{\text{RL}}(\theta)$, using gradient descent with respect to $\theta$. This loss is estimated using sampled trajectories, where each trajectory consists of a sequence of tuples of state, action, and reward.

\subsection{The LLM4Teach Framework}
The core principle of LLM4Teach involves the utilization of a pre-trained LLM as a teacher agent to guide a lightweight student RL agent in swiftly acquiring a policy for real-time decision-making to accomplish a specific embodied task. The student agent is allowed to interact with the environment and receive feedback from these interactions to rectify any errors provided by the teacher agent. Following the training phase, only the lightweight student agent is utilized during the testing phase, yet it owns superior capability in accomplishing the target task compared to its teacher. The conceptual framework of this approach is depicted in Figure~\ref{fig:framework}. %In the subsequent sections, we present the process of training the student agent in detail.

\subsection{On the LLM-based Teacher Agent}\label{sec:Interaction}
In accordance with \cite{ahn2022can}, we first notify the LLM of a set of $K$ option policies $\Pi: \{\pi_k : \mathcal{S} \rightarrow \mathcal{P}(\mathcal{A})\}$ related to the current task using appropriate prompts, where $k\in\{1,2,...,K\}$ denotes the option index. When presented with a state $s$, the student agent requests guidance from the teacher agent for the next step action. The teacher agent initially selects a high-level option $\pi_k$ from the set $\Pi$, prompted by a textual description $c(s)$ of the state $s$. Subsequently, an action suggestion $a \sim \pi_k(s)$ is generated based on the chosen option, serving as an instruction provided by the teacher.

\subsubsection{Generating Uncertainty-aware Instructions Using LLM}
The process of the student agent learning policies from the teacher agent can be seen as distilling important task-related knowledge from the LLM agent. As demonstrated in \cite{hinton2015distilling}, incorporating uncertainty into knowledge distillation can improve sample efficiency and prevent model over-fitting. Consequently, we propose having the LLM offer uncertainty-aware soft instructions to the student agent.
When the student agent sends a text description $c(s)$ to the teacher agent, the teacher agent responds by providing a soft decision $\pi_T(\cdot | s)$, i.e., a distribution over available options, in the following way:
\begin{equation}
    \pi_T(\cdot | s) = \sum_k Pr_{\text{LLM}}(k|c(s))\pi_k(\cdot | s),
    \label{eq:soft}
\end{equation}
where $Pr_{\text{LLM}}(k|c(s))$ represents the probability of the LLM teacher selecting the $k$th option given the textual description $c(s)$ of the current state $s$, and $\pi_k(\cdot | s)$ denotes the policy associated with the $k$th option. To estimate the uncertainties $Pr_{\text{LLM}}(k|c(s))$ in our experiments, we query the LLM multiple times with the same prompt to estimate the probability of each decision, similar to \cite{wang2022self}. An alternative approach is to access the logits of tokens relevant to option plans and convert them into probabilities \citep{carta2023grounding,ahn2022can}. We conduct an ablation study on these two approaches in subsection~\ref{sec: ablation}.

\begin{algorithm}[t]
    \caption{The student agent's policy learning algorithm}
    \label{alg: ours}
    \textbf{Require:} an LLM agent, pre-trained option policies~$\{\pi_k\}$, initial policy parameter value $\theta$, maximum allowable number of iterations $T$
    \begin{algorithmic}[1]
        \FOR {$i=1,2,...,T$}
            \STATE Collecting rollouts following the student agent's initial policy and stores the data in a buffer $\mathcal{D}$
            \FOR {each transition $(s,a,r)\in \mathcal{D}$}
                \STATE Generate a prompt with a textual description $c(s)$ of the state $s$ for the LLM-based teacher agent
                \STATE Get the soft decision of the LLM-based teacher agent according to Equation (\ref{eq:soft})
            \ENDFOR
        %\STATE $\lambda_i = \text{scheduler}(i)$
        %\Comment {Update Kickstarting coefficient following Eq.~\ref{eq: coeff}}
        \FOR {each gradient descent step}
            \STATE $\theta \leftarrow \theta - \alpha \nabla_\theta(\mathcal{L}_{\text{RL}}(\theta) + \lambda_i\mathbb{E}_{s}\mathcal{H}\left(\pi_T(\cdot|s)|| \pi_\theta(\cdot|s)\right))$
            %\Comment{Update $\theta$ following Eq.~\ref{eq: ks}}
        \ENDFOR
        \ENDFOR
    \end{algorithmic}
\end{algorithm}

\subsection{On the Learning Process of the Student Agent}
\label{sec: learning}
The learning process is summarized in Algorithm \ref{alg: ours}.
The policy of the student agent, denoted as $\pi_\theta(\cdot|s)$, is learned by minimizing the following loss function:
\begin{equation}
    \mathcal{L}(\theta) = \mathcal{L}_{\text{RL}}(\theta) + \lambda\mathbb{E}_{s\sim\pi_{\theta}}\mathcal{H}\left(\pi_T(\cdot|s)|| \pi_\theta(\cdot|s)\right),
    \label{eq: ks}
\end{equation}
where $\mathcal{L}_{\text{RL}}(\theta)$ denotes the traditional loss used in RL algorithms to encode the feedback from the environment. This loss is typically designed to maximize the expected return or rewards obtained by the agent. We incorporate the teacher agent's guidance into the student agent's learning process by introducing the regularization term $\mathcal{H}\left(\pi_T(\cdot|s)|| \pi_\theta(\cdot|s)\right)$ that describes the difference between the teacher policy $\pi_T(\cdot|s)$ and the student policy. This term captures the Kullback-Leibler (KL) divergence or Wasserstein distance between those two policies. To control the extent to which the student agent relies on the teacher agent, we introduce an annealing parameter $\lambda$. When $\lambda$ is set to zero, the learning process of the student agent reduces to a standard RL process without any influence from the teacher agent.

We initialize the annealing parameter $\lambda$ with larger values during the initial stages of training. This setup ensures that the student agent pays more attention to the guidance provided by the LLM-based teacher agent, aiming to align its policy with that of the teacher.
As the training progresses, we gradually decay $\lambda$, allowing the student agent to shift its focus towards maximizing its expected return. By reducing the influence of the teacher's guidance, the student agent becomes more independent in its decision-making process and emphasizes its own learned policy.
Specially, the annealing schedule used is designed as follow:
\begin{equation}
\lambda_i = \left\{
\begin{array}{ll}
    \lambda_0 - ki & \text{if}\quad i<i_1 \\
    \lambda_c & \text{if}\quad i_1<i<i_2 \\
    0 & \text{otherwise}
\end{array}, \right.
\label{eq: coeff}
\end{equation}
where $i$ represents the index of the training iteration, $k$ represents the decay rate, $\lambda_0$ is the initial value of $\lambda$, $\lambda_c$ is a constant value smaller than $\lambda_0$, which is maintained from the $i_1$th iteration to the $i_2$th iteration, $i_2$ indicates the point at which the connection to the LLM-based teacher agent is closed. For more details on the annealing schedule used in our experiments, see Appendix A.7.

This linear reduction of $\lambda$ enables a smooth transition for the student agent from heavily relying on the teacher's guidance to prioritizing the RL objective. It provides a balance between learning from the teacher and acquiring autonomy in decision-making. When $\lambda$ eventually reaches 0, we effectively remove the influence of the teacher's instructions on the student agent's policy, then the student agent no longer requires the teacher's guidance.

\section{Experiments}

We validated the performance of our method, LLM4Teach, through extensive experiments. The aim of the experiments is to demonstrate the specific advantages of LLM4Teach compared to RL baseline methods and approaches that solely rely on LLM for decision-making, and to test its potential in handling real-world sequential decision making problems.

\subsubsection{Simulation Platforms}
\paragraph{MiniGrid} offers a customizable grid world environment with various sizes, object types, and objectives, making it a simple representation of grid-based tasks \citep{MinigridMiniworld23}. These tasks pose a challenge for RL methods because of their sparse rewards.

\paragraph{Habitat} is a simulation platform specifically created to support the development of embodied AI systems \citep{szot2021habitat}. It provides a comprehensive framework for defining and executing various embodied AI tasks, such as navigation, object rearrangement, and question-answering. Additionally, Habitat enables detailed configuration of embodied agents, including their physical attributes and sensor specifications.

\subsubsection{Baseline Methods}
In the experiments, we include three baseline approaches to assess the performance of LLM4Teach.

\paragraph{LLM soly} operates in two stages. First, the LLM selects an option from a set of available pre-trained options. Then, a low-level action is generated following the selected option policy. In this configuration, only an LLM-based agent is utilized to make real-time decisions, without the involvement of the student agent. This approach allows us to investigate the potential of our proposed LLM4Teach framework in enabling the student agent to outperform its teacher in completing the desired task.

\paragraph{Hierarchical RL} is an RL baseline approach that involves training the student agent with pre-trained option policies \citep{matthews2022hierarchical}. In light of the hierarchical nature of the tasks, we explore such hierarchical RL approach in our experiments, so that we can assess the benefits of knowledge distillation from a pre-trained LLM that captures world knowledge.

\paragraph{Baseline RL} is a Tabula rasa RL baseline trained from scratch using the proximal policy optimization (PPO) algorithm \citep{schulman2017proximal}. The policy model structure and the training loss function are set the same as our student agent in LLM4Teach.

\subsection{Experiments on MiniGrid}
\label{sec:minigrid_env}

\subsubsection{Experimental Setting}
We created 4 procedurally generated tasks in the MiniGrid environment: \{\textit{SimpleDoorKey}, \textit{ColoredDoorKey}, \textit{LavaDoorKey} and \textit{DivergedDoorKey}\}. In each task, the agents are situated in rooms with varying layouts and their goal is to unlock the exit door using the correct key. In \textit{SimpleDoorKey}, the agent must explore the room, find a key, and use it to unlock the exit door. In \textit{ColoredDoorKey}, the exit door can only be unlocked with a key that matches its color, adding complexity for the agent to understand task-specific rules. \textit{LavaDoorKey} introduces hazard grids (Lava) to the room, requiring the agent to quickly adapt to new elements. \textit{DivergedDoorKey} presents two exit doors instead of one, allowing the agent to choose either door to escape, emphasizing the importance of using uncertainty-aware instructions to improve overall sample efficiency.

For every task, we incorporate 5 specialized options, which are: \{\textit{explore}, \textit{go to}, \textit{pickup}, \textit{drop}, \textit{open}\}. All options, with the exception of \textit{explore}, are dependent on specific conditions, such as interacting with an object, for example, \textit{pickup the red key}. These expert policies are compiled under the fundamental task of \textit{SimpleDoorKey}. Each option policy produces a Dirac delta distribution over actions based on the state. Additional information about the environments and options can be found in Appendix A.4.

\begin{figure}[t]
    \centering
    \includegraphics[width= 0.48\textwidth]{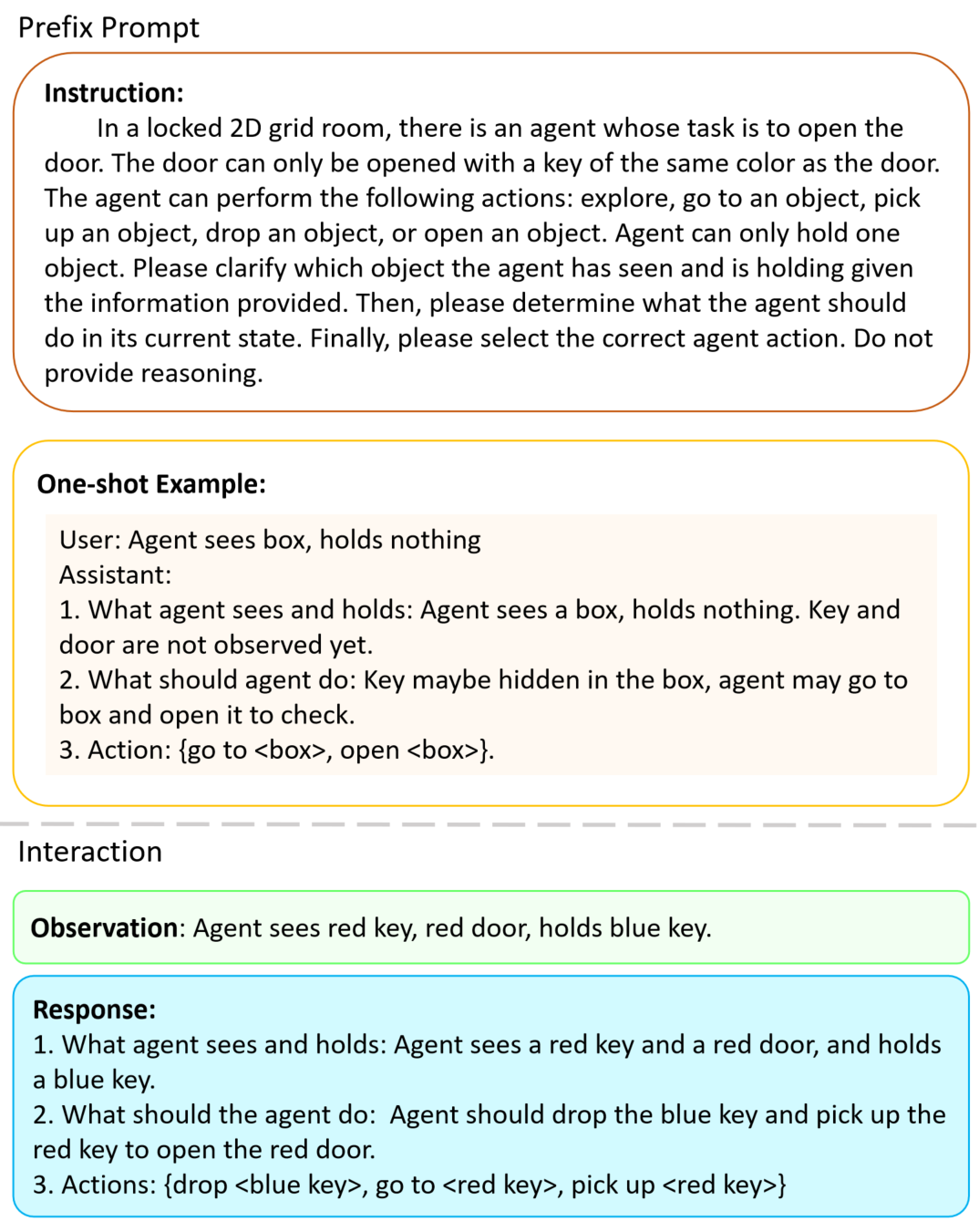}
    \caption{An example of a prefix prompt and an interaction between the student agent and the LLM-based teacher agent for the task \textit{ColoredDoorKey}. The Prefix prompt consists of two blocks: the instruction block briefly introduces the target problem and the CoT reasoning process; and the example block provides one arbitrary example of the expected format of the response from the LLM.}
    \label{fig:prompt_example}
\end{figure}

We use ChatGLM-turbo \citep{du2022glm} as the LLM to construct our teacher agent. This powerful model enables our teacher agent to possess complex reasoning capabilities. To leverage these capabilities, we employ Chain-of-thought (CoT) \citep{wei2022chain} style prompts. The CoT prompts consist of multiple stages that guide the LLM's decision-making process. Firstly, the LLM is prompted to summarize the scene, providing a condensed description of the environment. Secondly, it is instructed to reason about the appropriate course of action based on the given context. Finally, the LLM outputs its decision for the given task. To aid the LLM in understanding the reasoning process and ensuring correct output formatting, an arbitrary example is included in the prompt. This example serves as a reference point and helps the LLM grasp the desired output structure.
Figure~\ref{fig:prompt_example} illustrates an example of the dialogues generated by the LLM using this prompt setup in the \textit{ColoredDoorKey} task.

\subsubsection{Results on MiniGrid}
\begin{figure*}[t]
    \centering
    \includegraphics[width=0.9\textwidth]{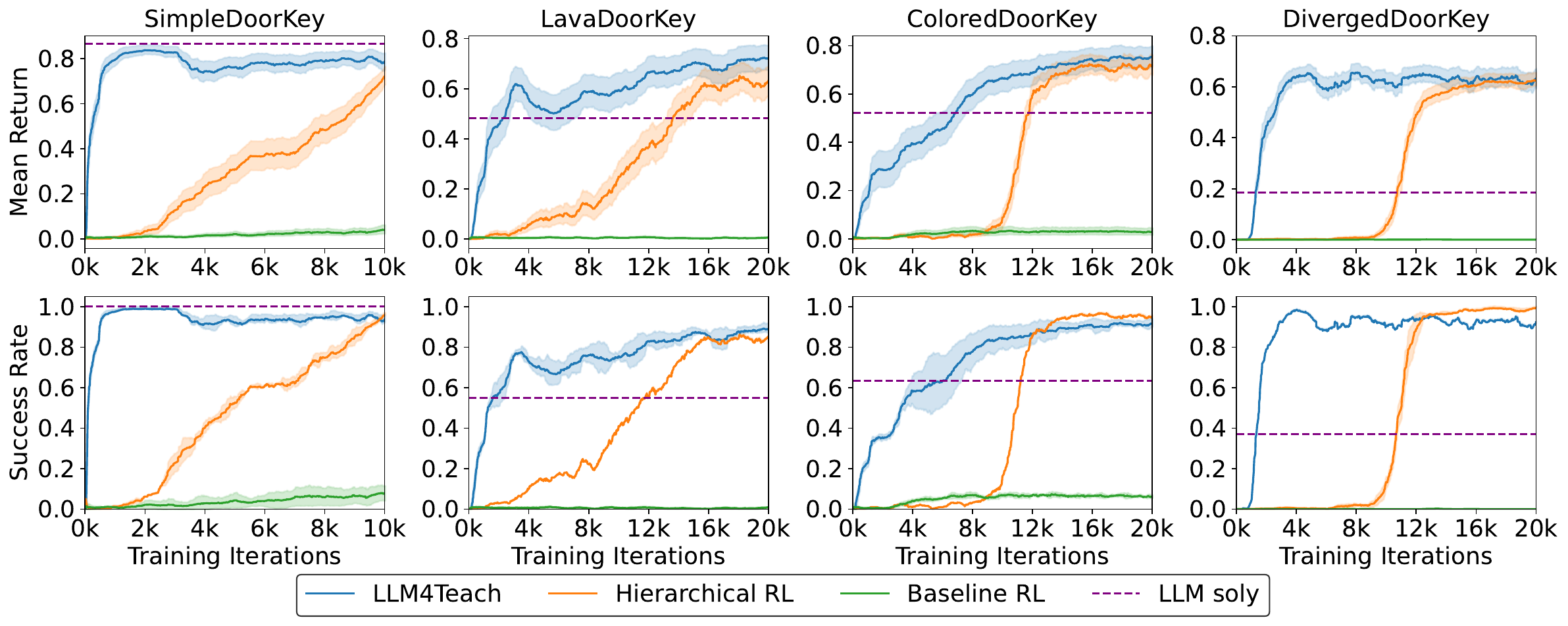}
    \caption{The tested average returns (top row) and task completion success rates (bottom row) vs. the training iteration index of the compared methods across four environments. The dotted vertical line indicates the point at which the teacher's guidance is diminished, i.e., when $\lambda_i = 0$. \textit{LLM soly} does not involve any learning, hence we report its average performance over 500 testing seeds, represented by a dashed horizontal line. For other approaches, we evaluate their policies every 10 iterations with 10 randomly generated testing seeds and report the averaged testing performance here. With our approach, the student agent effectively leverages the knowledge of the LLM-based teacher to bootstrap the early learning stage. Except for the \textit{SimpleDoorKey} task, the student agent in LLM4Teach ultimately outperforms the LLM-based agent by learning from environment feedback through minimizing a traditional RL loss.}
    \label{fig: curves}
\end{figure*} 
The main results in Figure~\ref{fig: curves} show that the baseline RL struggles to complete tasks, even the simplest one (\textit{SimpleDoorKey}), due to highly sparse rewards. In contrast, hierarchical RL eventually succeeds in the tasks but requires over 10,000 training iterations across all tasks. However, LLM4Teach, guided by the LLM-based teacher, effectively leverages the world knowledge embedded in the LLM, leading to significantly higher sample efficiency compared to prior art RL baselines with sparse rewards.

Results also show that LLM4Teach outperforms \textit{LLM soly} in terms of accumulated returns for all tasks, except for \textit{SimpleDoorKey}. \textit{SimpleDoorKey} is the simplest one, with low reasoning difficulty for LLM. Moreover, all option policies are designed based on this environment, so there is no issue of option policy transfer. Therefore, \textit{LLM soly} can achieve a success rate of nearly 100\% for the task.

For the other tasks which are more complex than \textit{SimpleDoorKey}, \textit{LLM soly} performs unsatisfactorily due to the lack of enough task-grounding knowledge. In comparison, LLM4Teach allows the student agent to learn task-grounding knowledge from the environmental feedback, thus performs much better than \textit{LLM soly}. For example, in \textit{ColoredDoorKey}, given the observation ``\textit{Agent sees a red key, a blue key, a blue door.}'', an LLM can suggest ``\textit{pickup the red key}'', while the right option is ``\textit{pickup the blue key}'', since only the the key with the same color of the door can be used to unlock the door. As a result, \textit{LLM soly} only achieves an average return of 0.52. In contrast, utilizing the student agent within LLM4Teach leads to a significantly higher average return of 0.77, as illustrated in Figure~\ref{fig: curves}. This is due to the student agent's ability to rectify its teacher's errors and adjust its behavior according to environmental feedback.

We have identified three major categories for the error policies generated by the LLM:
\begin{itemize}
    \item \textbf{Incorrect policies}, which are executable but result in task failure. For example, an incorrect policy could involve moving into the lava, leading to the failure of task completion.
    \item \textbf{Inefficient policies}, which are executable but not necessary for task completion. They can increase the number of steps required to accomplish the task, potentially resulting in time-out errors. For instance, an inefficient policy could involve continuously exploring even after finding the correct key and door, instead of directly proceeding to the door.
    \item \textbf{Inconsistent policies}, which are not executable due to non-compliance with behavioral logic or contextual constraints, e.g., attempting to pick up a new key without first dropping the key that the agent is currently holding.
\end{itemize}

\subsubsection{Ablation Study on Uncertainty-aware Instructions}
\label{sec: ablation}

\begin{figure}[t]
    \includegraphics[width=0.49\textwidth]{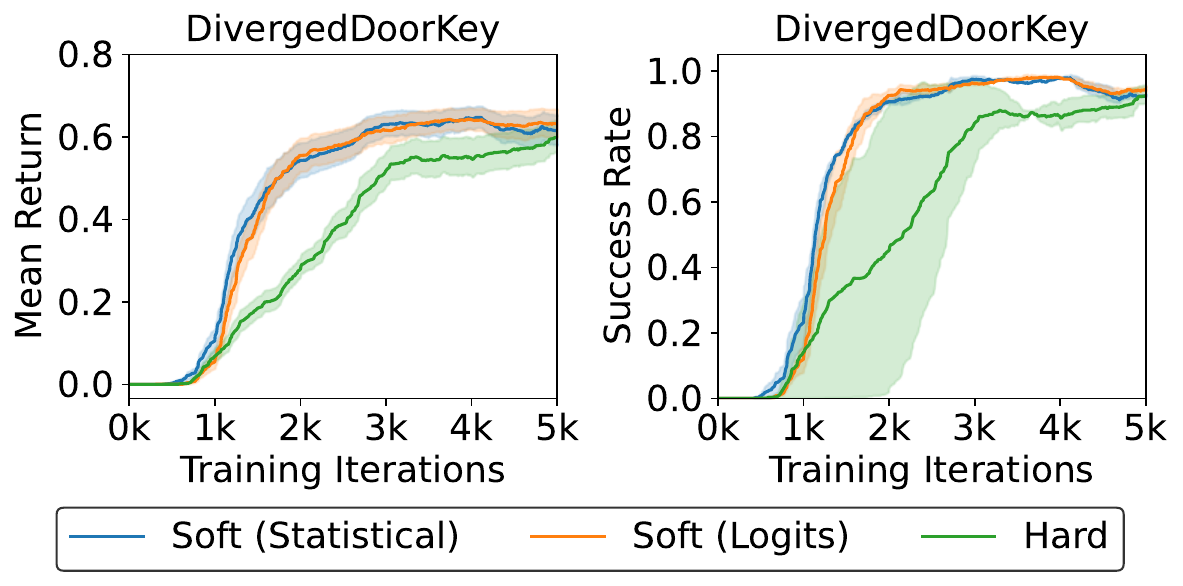}
    \caption{Ablation study on uncertainty-aware instructions. It shows that two types of uncertainty-aware instructions by the teacher both lead to improved sample efficiency for the student agent.}
    \label{fig: ablation}
\end{figure}

As presented in subsection \ref{sec:Interaction}, the teacher agent in LLM4Teach offers uncertainty-aware instructions to the student agent, which is a distinguishing feature compared to previous LLM-based agents (e.g., in \cite{ahn2022can}), where deterministic feedback is provided upon receiving a query.

We conducted ablation studies to investigate the benefits of using uncertainty-aware instructions instead of deterministic ones in the \textit{DivergedDoorKey} task. We considered two approaches for the LLM to provide uncertainty-aware soft instructions. The first one is to query the LLM multiple times with the same prompt to \textbf{statistically} estimate the probability of each decision, similar to \cite{wang2022self}.
The other approach is to access the \textbf{logits} of tokens relevant to option plans and convert them into probabilities \citep{carta2023grounding,ahn2022can}.
We compare these two approaches with a \textbf{hard} instruction baseline, where the LLM's responses are directly used as deterministic instructions.

The result of the ablation study is shown in Figure~\ref{fig: ablation}. It is shown that utilizing uncertainty-aware instructions improves the overall sample efficiency compared to using deterministic ones. Moreover, there is no significant disparity in performance between the two approaches for generating uncertainty-aware instructions. The first approach is simpler to implement in practical scenarios but consumes more computational resources due to multiple queries to LLMs, particularly when the observation space is large. The second approach necessitates access to logits, making it applicable only to open-source LLMs.

\subsection{Experiments on Habitat}
\label{sec:habitat_env}

To evaluate the potential applicability of our method in real-world scenarios, we conducted additional experiments using Habitat \citep{szot2021habitat}, which involves continuous action spaces and
high-dimensional observations.

\subsubsection{Experimental Setting}

\begin{figure}[t]
    \centering
    \begin{subfigure}{0.18\textwidth}
        \includegraphics[width= \textwidth]{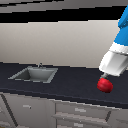}
    \end{subfigure}
    \begin{subfigure}{0.18\textwidth}
        \includegraphics[width= \textwidth]{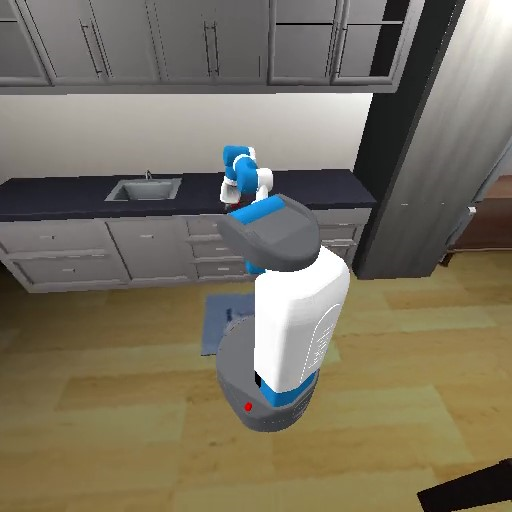}
    \end{subfigure}
    \caption{Habitat environment. Left: The visual observation from the onboard camera. Right: A view of the acting robot and its workspace from a third-party camera. Note that the third-party camera mentioned is purely for illustrative purposes and is not utilized during either the training or testing phases.}
    \label{fig: habitat}
\end{figure}

We focus on a manipulation task called \textit{Nav \& Pick}. The objective of the robotic agent is to navigate to the table without any collisions and subsequently perform a precise object pickup. Refer to Figure~\ref{fig: habitat} for a visual representation.

We conduct separate pre-training for two high-level options, namely \textit{Navigate} and \textit{Pick}. These options are utilized by both LLM4Teach and the hierarchical RL baseline \citep{matthews2022hierarchical}. To ensure the effectiveness of option training, we employ ten distinct training environment specifications, each with varying object and target locations. Furthermore, the agent's initial positions are randomly generated upon environment reset, ensuring diverse training scenarios. For each option, we utilize a ResNet18 backbone in conjunction with a 2-layer MLP architecture to train the corresponding models. For more detailed information about the environments and training parameters, refer to Appendix A.5.

We select the Vicuna-7b model \citep{vicuna2023} as the LLM used in LLM4Teach, following a similar prompt design as in previous experiments on Minigrid. We utilize visual observations captured by the on-board camera as input queries for the LLM. To enable the LLM-based teacher agent to comprehend these visual inputs, we utilize a pre-configured translator in the same way as in \cite{hu2024enable} to generate natural language descriptions which list identified objects in the visual inputs. %Alternatively, pre-trained visual-language models such as CLIP \citep{radford2021learning} can be utilized for this purpose.

\subsubsection{Results on Habitat}
\begin{figure}[t]
    \centering
    \includegraphics[width=0.45\textwidth]{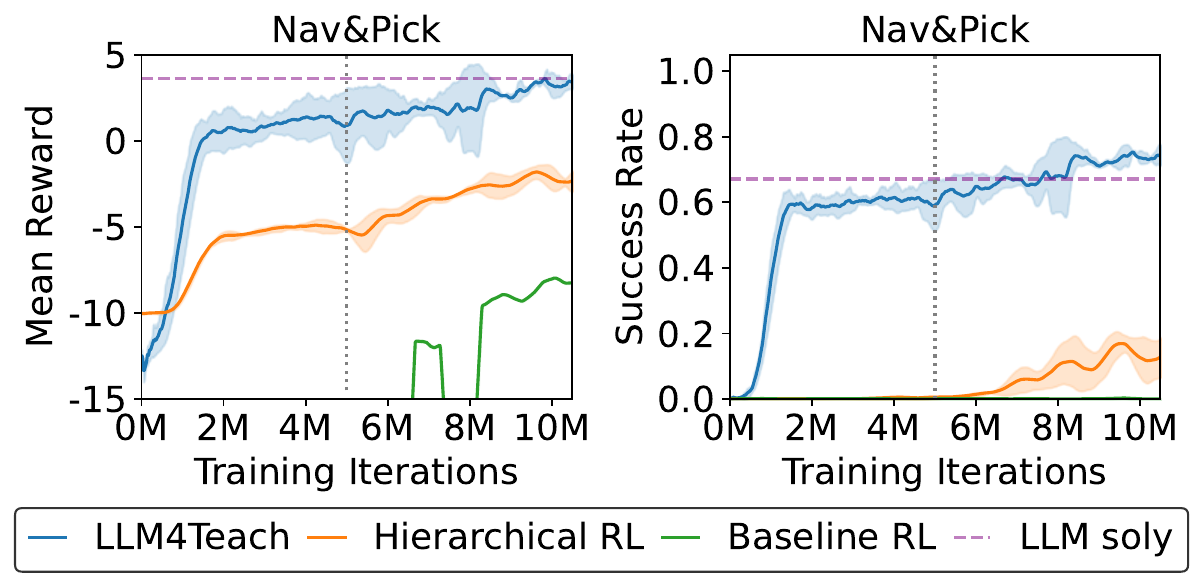}
    \caption{The tested average returns (left) and task completion success rates (right) vs. the training iteration index of the compared methods on the \textit{Nav\&Pick} task. For explanations of the lines and curves in the figure, see the caption of Figure \ref{fig: curves}.}
    \label{fig: habitat_result}
\end{figure}

Due to the task being limited to home scenarios, the LLM effectively covers the common-sense reasoning abilities required to successfully complete the task. This results in few erroneous decision-making during option selection. Consequently, the task completion rate and average returns for \textit{LLM soly}, as depicted in Figure~\ref{fig: habitat_result}, are relatively high. In contrast, the RL baselines struggle to complete the task due to the scarcity of rewards. Our approach, LLM4Teach, consistently outperforms all RL-based baselines in terms of both sample efficiency and asymptotic performance. This highlights the effective utilization of the LLM-based teacher's knowledge by the student agent in LLM4Teach, facilitating the learning of appropriate policies. Given enough training iterations, our approach exhibits a higher success rate compared to \textit{LLM soly}. The primary advantage of LLM4Teach is that it is an extremely lightweight RL-based student agent specifically designed for utilization in the final online testing phase.

\section{Concluding Remarks}

Both RL and LLMs have limitations in handling complex sequential decision-making problems. RL often lacks sample efficiency and incurs high exploration costs, while LLMs are prone to decision errors and have high deployment costs. Combining LLMs with RL to overcome these limitations is a natural idea, but creating an effective interface between them poses challenges. LLMs utilize texts as input and output, making them suitable for providing high-level instructions, whereas RL operates at a much lower fine-grained level and uses numerical vectors instead of texts as inputs and outputs.

Here we present LLM4Teach, a novel framework that combines LLMs and RL for embodied sequential decision-making tasks. Our approach leverages the reasoning capabilities of LLMs to develop a highly capable RL-based student agent. In particular, we use the LLM to provide high-level suggestions on available options for policy training of the student agent. Extensive experiments demonstrate that our student agent outperforms all RL baselines in sample efficiency. Meanwhile, it achieves superior performance to \textit{LLM soly} in terms of task completion success rate with much fewer computational resources during online testing. For instance, in MiniGrid experiments, the student agent's model size is 24K compared to LLM's 130B. Similarly, in Habitat experiments, the student agent's model size is 10M while an LLM's is 7B.

\appendix

\section*{Ethical Statement}
There are no ethical issues.

\section*{Acknowledgments}
This work was supported by Exploratory Research Project (No.2022RC0AN02) of Zhejiang Lab.

\section*{Contribution Statement}
Z. Zhou, B. Hu, C. Zhao and P. Zhang contribute equally. B. Liu is project lead.

\bibliographystyle{named}
\bibliography{ijcai24}
\newpage
\appendix
\section{Appendix}
\subsection{How do we choose the $K$ options in correspondence to a specific task?}
We select $K$ options using prior knowledge, which is obtained from human experts or pre-trained LLMs. For instance, if we ask an LLM about the motion primitives associated with robots, it might respond as follows: ``there are several motion primitives related to robots, such as straight-line motion, arc motion, rotation motion, grasping motion, lifting motion, placing motion, swing motion, and elevating motion". Each motion primitive corresponds to an option in our framework. The robot/agent can utilize combinations of these options to accomplish various tasks. The motion primitives used in our experiments are ``explore", ``go to", ``pickup", ``drop", ``open", as outlined in subsection A.4.
\subsection{On the textual description $c$}
An illustrative example of the textual description $c$ is as follows: ``The agent observes a red key, a red door, and holds a blue key", as depicted in Figure 2. We have developed a parser to generate $c$.  The parser utilizes a predefined text template with a consistent structure and fills in all items that fall within the agent's field of view, including those held in its hands, at each time step.
\subsection{Is it expensive to query the LLM to obtain a good estimation of $Pr_{\text{LLM}}$?}
To make statistical estimations on decision uncertainties, if we choose to query LLMs multiple times using the same prompt, it can be costly. However, if we opt to use logits of the LLM, it is a more affordable approach, but it requires the LLM to be open-sourced, allowing us to access the logits. Our findings indicate that both approaches yield comparable results, as shown in Figure 4.
\subsection{Minigrid Experiments}
\label{sec: option}
\subsubsection{Option Framework}
We address the hierarchical structure of target tasks by employing an option framework \citep{sutton1999between}, which defines an option as a sub-policy that specifies a behavior extended over time. Each option $\omega$ is defined by the triplet $(\mathcal{I}, \pi, \beta)$, representing the set of initiation states, the acting policy, and the termination condition, respectively.
%In this section, we provide a detailed description of the option set used in the LLM soly baseline, as presented in Section \ref{sec:minigrid_env}. %We specifically focus on the termination conditions employed for each option.

For the MiniGrid environments, we utilize a set of options consisting of: \{\textit{explore}, \textit{go to}, \textit{pickup}, \textit{drop}, \textit{open}\}. Each option can be initiated from any state, i.e., $\mathcal{I}_\omega = \mathcal{S}$ for all options. %Additionally, each option has a maximum length of 100 steps, at which point the agent terminates the option.
Here is a breakdown of each option and its associated termination conditions:
\begin{itemize}
    \item \{\textit{explore}\}: During exploration, the agent systematically scans the unexplored grid row-by-row following a predetermined strategy. This option terminates when the agent observes walls forming a closed area.
    \item \{\textit{go to}\}: The agent plans a path to the target object using the $A^*$ algorithm and terminates the option upon reaching the target object.
    \item \{\textit{pickup}\}: The agent attempts to pick up the target object if it is not already holding another object. Otherwise, it first drops the current object at the nearest available position before picking up the new one.
    \item \{\textit{drop}\}: The agent drops the object it is currently holding at the nearest available position.
    \item \{\textit{open}\}: This is a one-step action that attempts to interact with the object in front of the agent.
\end{itemize}

\subsubsection{Hyperparameters}
\label{sec: appendix_ppo}
We choose Proximal Policy Optimization (PPO) \citep{schulman2017proximal} as the underlying RL algorithm for all methods in MiniGrid. The specific hyperparameters are listed in Table~\ref{table: ppo_hyperparameters} for reference.
\begin{table}[h]
    \centering
    \begin{tabular}{l | l  }
    Variable  & Value \\
      \hline
    Number of trajectories per iteration & 10\\
    Number of epochs per iteration & 3 \\
    Minibatch size & 128\\
    Entropy loss coefficient & 0.001 \\
    Value function loss coefficient & 0.5\\
    Discount factor & 0.99 \\
    Learning rate & 0.001 \\
    Clipping parameter & 0.2\\
    Maximum gradient norm & 0.5 \\
    \end{tabular}

    \caption{PPO hyperparameters in MiniGrid experiments.}
    \label{table: ppo_hyperparameters}

\end{table}

\subsection{Habitat Experiments}
\label{sec: hab_training}
%In this section, we provide a detailed description of the training setting in Section \ref{sec:habitat_env}.
\subsubsection{Task Details}
In our Habitat experiments, the robot agent is equipped with a wheeled base, a 7-degree-of-freedom (DoF) arm manipulator, and a parallel-jaw gripper. Additionally, it is equipped with a camera mounted on its ``head" that provides a $90^{\circ}$ field of view and captures visual data at a resolution of $256\times 256$ pixels. Therefore, the observation space of the environment consists of a visual observation denoted as $o_v \in \mathbb{R}^{256\times 256\times 1}$ from the depth camera. It also includes a sensor observation $o_s \in \mathbb{R}^{24}$ obtained from various sensors such as joint sensors, gripping sensors, the end effector of the arm, object, and target GPS sensors, among others. The action space in our setup is 11-dimensional, comprising 2 actions for controlling the robot positions, 7 actions for controlling the robot arm, 1 action indicating whether the robot is holding an object, and 1 action indicating termination. This action space enables the agent to execute precise movements and manipulations required to accomplish the target task. For detailed information on the training of option policies for the LLM agent, refer to the Habitat documentation \citep{szot2021habitat}. These option policies for the teacher model are kept fixed during the knowledge distillation process to ensure consistency and stability during execution.

The agent is trained using the following reward function:
\begin{align*}
 r_t = & 5\mathbb{I}_{pickup} + \Delta^{o}_{arm} \mathbb{I}_{!holding} - 10\mathbb{I}_{force} - 0.005,
\end{align*}
where $\mathbb{I}_{pickup}$ is an indicator function that is 1 if the agent has picked up the object, $\mathbb{I}_{holding}$ is an indicator function that is 1 if the robot is holding an object, $\Delta^{o}_{arm}$ represents the change in Euclidean distance between the end-effector and the object, and $\mathbb{I}_{force}$ is an indicator function that is 1 if the force on the robot due to collision exceeds a specified limit. Additionally, a slack reward of -0.005 is given to incentivize the agent to complete the task as quickly as possible.

\subsubsection{Hyperparameters}
\label{sec: appendix_ddppo}
We train the policies using Decentralized Distributed Proximal Policy Optimization (DD-PPO) \citep{wijmans2019dd} with Wasserstein distance regularization terms. The hyperparameters and their values used in the experiments are listed in Table~\ref{table: ddppo_hyperparameters}.
\begin{table}[h]
    \centering
    \begin{tabular}{l | l  }
    Variable  & Value \\
      \hline
    Number of environment & 10\\
    Number of epochs per iteration & 1 \\
    Minibatch size & 1024\\
    Entropy loss coefficient & 0.001 \\
    Value function loss coefficient & 0.5\\
    Discount factor & 0.99 \\
    GAE lambda & 0.95 \\
    Learning rate & 0.0003 \\
    Epsilon value & 0.00001 \\
    Clipping parameter & 0.2 \\
    Maximum gradient norm & 0.5 \\
    Initial annealing parameter $\lambda_0$ & 10 \\
    Maintain annealing parameter $\lambda_c$ & 0.1 \\
    Maintain annealing parameter iteration $i_1$ & 5e6 \\
    Remove annealing parameter iteration $i_2$ & 1e7
    \end{tabular}

    \caption{DD-PPO hyperparameters in Habitat experiments.}
    \label{table: ddppo_hyperparameters}
\end{table}
\begin{table*}[h]
    % \scalebox{0.7}{
    \centering
    \begin{tabular}{c | c | c c c }
    \toprule
    \multirow{2}{*}{Task} & \multirow{2}{*}{Method}  & \multicolumn{3}{c}{Performances} \\
    & & Avg. episode length & Avg. Return & Success Rate \\
      \midrule
    \multirow{4}{*}{\textit{SimpleDoorKey}} &
    LLM soly & \textbf{22.28} & \textbf{0.87} & \textbf{100.0\%} \\
    & Hierarchical RL & 49.83 & 0.70 & 96.0\% \\
    & Baseline RL & 143.99 & 0.05 & 10.1\% \\
    & LLM4Teach & 30.49 & 0.81 & 96.4\% \\
    \midrule
    \multirow{4}{*}{\textit{LavaDoorKey}} &
    LLM soly & \textbf{15.79} & 0.48 & 54.6\% \\
    & Hierarchical RL & 43.05 & 0.63 & 85.1\% \\
    & Baseline RL & 149.61 & 0.00 & 5.0\% \\
    & LLM4Teach & 36.84 & \textbf{0.75} & \textbf{90.9\%} \\
    \midrule
    \multirow{4}{*}{\textit{ColoredDoorKey}} &
    LLM soly & 90.73 & 0.40 & 46.8\% \\
    & Hierarchical RL & 42.69 & 0.74 & \textbf{95.8\%} \\
    & Baseline RL & 145.84 & 0.03 & 6.6\% \\
    & LLM4Teach & \textbf{36.27} & \textbf{0.78} & 94.0\% \\
    \midrule
    \multirow{4}{*}{\textit{DivergedDoorKey}} &
    LLM soly & 50.10 & 0.19 & 37.2\% \\
    & Hierarchical RL & \textbf{27.21} & \textbf{0.59} & \textbf{98.0\%} \\
    & Baseline RL & 59.63 & 0.01 & 1.9\% \\
    & LLM4Teach & 30.65 & 0.53 & 93.1\% \\
    \midrule
    \multirow{4}{*}{\textit{Nav \& Pick}} &
    LLM soly & 330.23 & 2.18 & 62\% \\
    & Hierarchical RL & 534.68 & -1.51 & 12\% \\
    & Baseline RL & 602.69 & -6.72 & 0\% \\
    & LLM4Teach & \textbf{233.45} & \textbf{2.39} & \textbf{69\%} \\
    \bottomrule
    \end{tabular}
    % }
    \caption{Summary of asymptotic performances of all methods on all tasks.}
    \label{table: SimpleDoorKey_result}
% \vspace{-0.5cm}
\end{table*} 
\subsection{Detailed Results}
\label{sec: appendix_results}
We provide the detailed asymptotic performances for all tasks in Table~\ref{table: SimpleDoorKey_result}. The Minigrid results are averaged over 1000 test runs, and the Habitat results are averaged over 100 test runs.

\subsection{Additional study on the annealing schedule}\label{sec:additional}
\begin{figure}[htb]
    \centering
    \begin{subfigure}{0.39 \textwidth}
        \includegraphics[width= \textwidth]{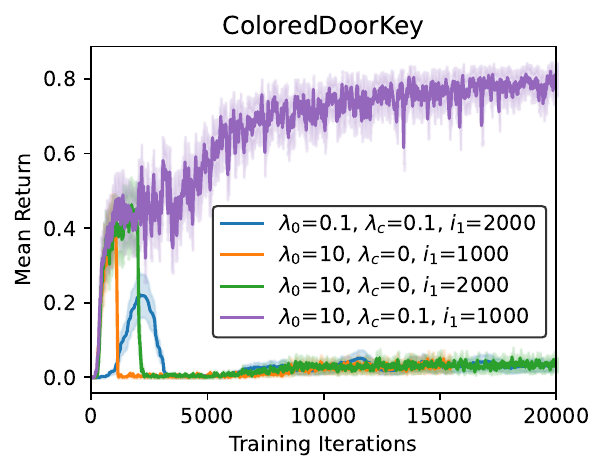}
    \end{subfigure}
    \begin{subfigure}{0.39 \textwidth}
        \includegraphics[width= \textwidth]{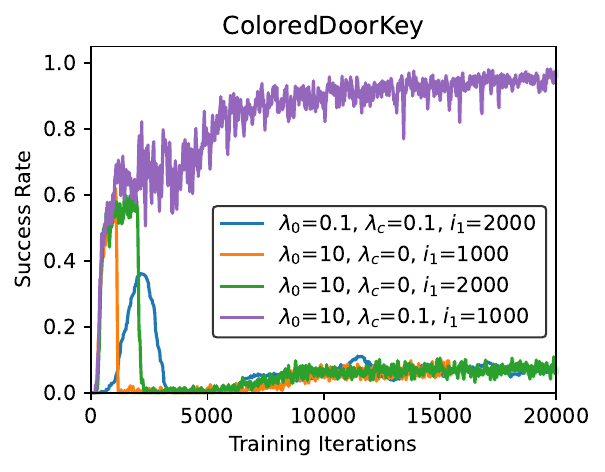}
    \end{subfigure}
    \caption{The training mean return (top) and success rates (bottom) of all compared annealing schedule designs in the \textit{ColoredDoorKey} environment}
    \label{fig: schedule}
    % \vspace{-2.0cm}
\end{figure}

As described in Section~\ref{sec: learning} (see Equation~\ref{eq: coeff}), the value of $\lambda$ depends on a set of hyper-parameters, including $\lambda_0, \lambda_c, i_1, i_2$, and $k$. In our experiments,
we set $i_2 = 2000$ for all MiniGrid tasks, meaning that the value of $\lambda$ reduces to 0 after 2000 iterations. In this section, we present an ablation study that compares different annealing schedules for $\lambda$. The schedules considered are as follows:
\begin{enumerate}
    \item Constant value: $\lambda$ takes a constant value of 0.1 then decreases to 0 at the $i_2$th iteration ($\lambda_0=0.1,\lambda_c=0.1, i_1=2000$);
    \item Linearly decaying value: the value of $\lambda$ linearly decays from 10 to 0 over the first 1000 iterations ($\lambda_0=10, \lambda_c=0, i_1=1000$) or 2000 iterations ($\lambda_0=10, \lambda_c=0, i_1=2000$);
    \item Stepwise value: The value of $\lambda$ linearly decreases from 10 to 0.1 over 1000 iterations, and then remains constant at 0.1 for some iterations before eventually reducing to 0 ($\lambda_0=10, \lambda_c=0.1, i_1=1000$).%($k=0.01,t=2000$, where $k$ is the decay rate and $t$ is the termination step).
\end{enumerate}
As depicted in Figure~\ref{fig: schedule}, we observed that only the last annealing schedule yielded successful results. We argue that this result is attributed to the fact that when the LLM-based teacher is removed, the student agent shall experience a period of policy oscillation in the short term. By allowing the student agent to adapt to the subtle influence of the LLM-based teacher initially and gradually removing it, the agent can navigate through this learning period more smoothly. Empirically we suggest setting $i_1$ as the iteration number when the regularization term $\mathcal{H}\left(\pi_T(\cdot|s)|| \pi_\theta(\cdot|s)\right)$ converges, and setting $i_2$ as two times $i_1$.
These findings highlight the importance of carefully choosing an appropriate annealing schedule for $\lambda$ to ensure effective knowledge transfer and smooth adaptation of the student agent during the training process. 

To summarize, we set parameter values of the annealing schedule in an empirical way. Specifically, we simulate the learning process of a child receiving guidance from a teacher. In the early stages of learning, the child has no prior knowledge, so we let it rely heavily on the teacher's demonstrations. As the learning progresses, we gradually reduce its dependence on the teacher until it reaches zero. Theoretical research on annealing schedules design for simulated annealing based optimization and stochastic gradient descent based deep neural network training may provide theoretical guidance, as their underlying mechanisms share similarities at a conceptual level.
\subsection{``LLM as a policy" vs. using ``LLM as a critic"}    
We argue that using ``LLM as a policy" (as in this work) vs. using ``LLM as a critic", such as the work presented in \cite{kwon2023reward}, are two complemented ways of training RL using the prior knowledge encoded in the LLM. The latter primarily focuses on addressing the challenge of designing rewards in scenarios where it is difficult; the former aims to solve the problem of high sample complexity during RL training. In real-world scenarios, if the main difficulty lies in reward design, the latter may be more suitable. However, if the focus is on improving the sample efficiency of RL training rather than reward design, our method proposed here can be considered. From a fundamental conceptual perspective, we can also compare the two approaches. In the former approach, we have the LLM teacher personally demonstrate how to make decisions given a specific observation. In the latter approach, the LLM teacher only provides evaluations of the student's behaviors without personally demonstrating it. We argue that, if the teacher is highly qualified, the approach of personal demonstration is more likely to result in higher training efficiency. This has already been demonstrated in some imitation learning-related literature, see e.g., in \cite{ramirez2022model}. As we know, in reality, humans can learn from both demonstrations and rewards. So we guess it is possible to combine these two approaches, using LLM as both a policy and a critic, to efficiently solve complex tasks.

\end{document}